\documentclass[10pt,twocolumn,letterpaper]{article}

\usepackage{iccv}
\usepackage{times}
\usepackage{epsfig}
\usepackage{graphicx}
\usepackage{amsmath}
\usepackage{amssymb}

\usepackage{url}
\usepackage{enumitem}
\usepackage{braket}
\usepackage{ulem}
\usepackage{pifont}
\usepackage{algorithmic}
\usepackage{subfigure}
\usepackage{multirow}
\usepackage{makecell}
\usepackage{balance}
\usepackage{booktabs}
\usepackage[table,xcdraw]{xcolor}
\usepackage{color}

\usepackage[pagebackref=true,breaklinks=true,letterpaper=true,colorlinks,bookmarks=false]{hyperref}

\iccvfinalcopy 

\def\iccvPaperID{2622} 

\ificcvfinal\fi

\begin{document}

\title{MixSpeech: Cross-Modality Self-Learning with Audio-Visual Stream Mixup\\
for Visual Speech Translation and Recognition}


\author{
Xize Cheng,
Linjun Li,
Tao Jin,
Rongjie Huang,
Wang Lin,\\
Zehan Wang,
Huangdai Liu,
Ye Wang,
Aoxiong Yin,
Zhou Zhao
\\
Zhejiang University\\
\tt\small \{chengxize,lilinjun21,jint\_zju,rongjiehuang,linwanglw\}@zju.edu.cn\\
\tt\small \{wangzehan01,liuhuadai,22151150,yinaoxiong,zhaozhou\}@zju.edu.cn\\
}

\maketitle
\ificcvfinal\thispagestyle{empty}\fi

\begin{abstract}


Multi-media communications facilitate global interaction among people. However, despite researchers exploring cross-lingual translation techniques such as machine translation and audio speech translation to overcome language barriers, there is still a shortage of cross-lingual studies on visual speech. This lack of research is mainly due to the absence of datasets containing visual speech and translated text pairs. In this paper, we present \textbf{AVMuST-TED}, the first dataset for \textbf{A}udio-\textbf{V}isual \textbf{Mu}ltilingual \textbf{S}peech \textbf{T}ranslation, derived from \textbf{TED} talks. Nonetheless, visual speech is not as distinguishable as audio speech, making it difficult to develop a mapping from source speech phonemes to the target language text. To address this issue, we propose MixSpeech, a cross-modality self-learning framework that utilizes audio speech to regularize the training of visual speech tasks. To further minimize the cross-modality gap and its impact on knowledge transfer, we suggest adopting mixed speech, which is created by interpolating audio and visual streams, along with a curriculum learning strategy to adjust the mixing ratio as needed. MixSpeech enhances speech translation in noisy environments, improving BLEU scores for four languages on AVMuST-TED by +1.4 to +4.2. Moreover, it achieves state-of-the-art performance in lip reading on CMLR (11.1\%), LRS2 (25.5\%), and LRS3 (28.0\%).
\end{abstract}


\begin{figure}[th]
    \centering
    \includegraphics[scale=0.46]{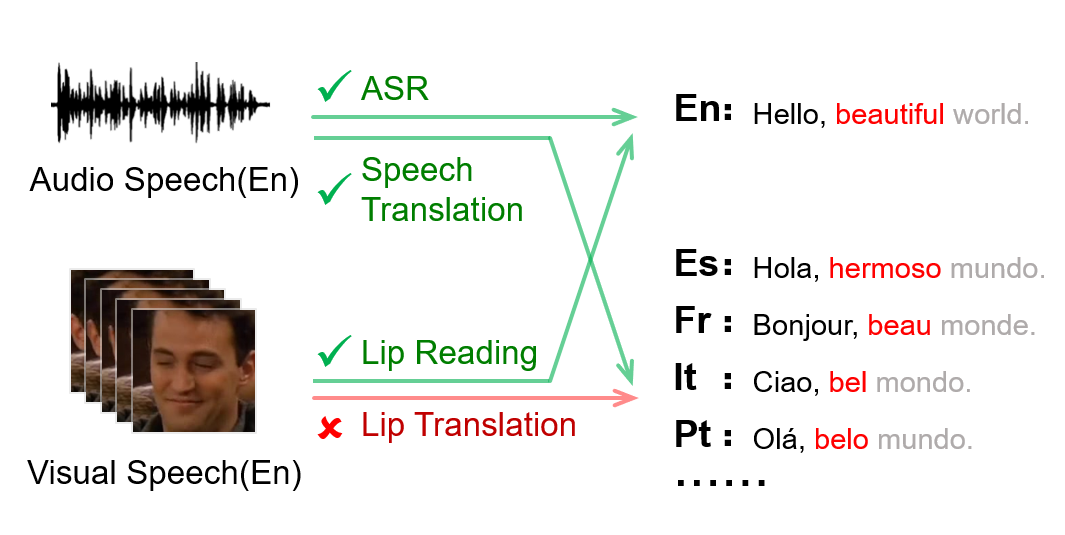}
    \caption{Diagram of speech tasks. Audio speech and visual speech are paired parallel speech streams which can be employed for speech recognition and speech translation. However, only Lip-Translation remains unexplored.}
    \label{fig:intro}
\end{figure}

\section{Introduction}
Multi-media techniques, including Audio-Visual Speech Recognition (AVSR) \cite{assael2016lipnet,afouras2018deep,afouras2018lrs3,shi2022learning}, Audio-Visual Speech Translation (AVST) \cite{berard2016listen,liu2019end,wang2022discrete}, and Audio-Visual Speech Generation (AVSG) \cite{prajwal2020lip,lahiri2021lipsync3d,hu2021neural}, are commonly employed in various online communication scenarios, such as conferences, education, and healthcare. As a tool for ultra-remote communication, many online interactions involve multiple languages, prompting the need for addressing cross-lingual challenges. Several works have attempted to tackle these challenges, including Machine Translation (MT) \cite{brown1990statistical,liu2020multilingual,costa2022no} for text utterance, Speech Translation (ST) \cite{sperber2017neural,fang2022stemm} for audio utterance, and Speech-to-Speech Translation (S2ST) \cite{sperber2017neural,fang2022stemm,dong2022leveraging,lee2021direct,huang2022transpeech} for simultaneous interpretation.
However, research on cross-lingual visual speech is still limited, as illustrated in Figure \ref{fig:intro}.
As an essential component of multi-media speech, visual speech can be combined with audio to enhance the recognition and understanding of speech content as audio-visual speech \cite{afouras2018deep,afouras2018lrs3,shi2022robust}, and is the unique resource for speech content understanding in audio-disabled scenarios \cite{lin2021simullr}.

Visual speech translation has never been studied, mainly for the lack of visual speech datasets with translated texts in different languages.
The few remaining works \cite{song2022talking,waibel2022face,nvidia2022} also cannot be quantitatively verified for this reason, making them unconvincing.
The available visual speech corpus is often very scarce compared to audio speech owing to the high demands of visual speech for model training, which requires mostly-frontal and high-resolution videos with a sufficiently high frame rate, such that motions around the lip area are clearly captured \cite{hsu2022single}. In this paper, we propose the first Audio-Visual Multilingual Speech Translation dataset, AVMuST-TED. During the process of acquisition, we first screen out videos with professional translations in four different languages from TED talk which performs strict translation and review processes, and then determine the real speaker's talking head by checking whether each pair of visual speech (\ie, talking head) and audio speech matches in the manner of \cite{afouras2018deep,afouras2018lrs3}. Incidentally, this dataset can also be used for quantitative evaluation of other multi-modality translation tasks, such as cross-lingua audio-visual speech generation \cite{ritter1999face,waibel2022face}. 


The cascaded model comprising of a speech recognition model and a machine translation model can handle speech translation tasks but suffers from error accumulation due to model cascades and cannot process languages without text (\eg, Minnan). Our proposed end-to-end model, which can translate directly from source speech to target text, addresses the above issues. However, visual speech is less distinguishable than audio speech, making it difficult to develop a mapping from source speech phonemes to the target language text. To address this, we introduce MixSpeech, a method that first pretrains the decoder using high-discrimination audio speech to obtain a mapping from speech phonemes to text and then generalizes this mapping to the visual speech task through cross-modality self-learning. Furthermore, since audio speech and visual speech are two distinct modalities of speech, there is a significant modality gap between them that hinders knowledge transfer. To narrow this gap and improve knowledge transfer, we propose mixed speech, which is created by interpolating audio and visual streams, rather than relying solely on audio speech. We also propose a curriculum-learning \cite{bengio2009curriculum} based strategy to adjust the mixing ratio as the training progresses and cross-modality integration deepens.



The code and dataset are available\footnote{\url{https://github.com/Exgc/AVMuST-TED}}, the main contributions of this paper are as follows:
\begin{itemize}
    \setlength{\leftmargin}{0pt}
    \setlength{\itemindent}{0pt}
    \setlength{\tabcolsep}{0pt}
    \setlength{\parskip}{0pt}
    \setlength{\partopsep}{0pt}
    \setlength{\itemsep}{0pt}
    \setlength{\topsep}{0pt}
    \setlength{\parsep}{0pt}
    \item We present the first lip-translation baseline and introduce the Audio-Visual Multilingual Speech Translation dataset, AVMuST-TED.
    
    

    \item We present a cross-modality self-learning framework that leverages distinguishable audio speech translation to regularize visual speech translation for effective cross-modality knowledge transfer.

    \item We present to adopt the mixed speech, interpolated from audio and visual speeches, and a curriculum-learning based mixing ratio adjustment strategy to reduce the inter-modality gap during knowledge transfer.


    \item We achieve state-of-the-art performance in lip translation for four languages on AVMuST-TED, with a +1.4 to +4.2 boost in BLEU scores and in lip reading on CMLR (11.1\%), LRS2 (25.5\%) and LRS3 (28.0\%).
\end{itemize}

\section{Related work}

\subsection{Audio-Visual Speech}
Audio and visual speeches are two separate modalities that convey speech content. Numerous works \cite{panayotov2015librispeech,chung2017out,afouras2018deep,afouras2018lrs3,prajwal2022sub,huang2022fastdiff,huang2022prodiff} have explored ways to extract information from speech using these modalities. Speech recognition \cite{panayotov2015librispeech,baevski2020wav2vec,hsu2021hubert} is widely used in online meetings and social applications to recognize speech content. Speech translation \cite{sperber2017neural,ye2021end,fang2022stemm} is commonly used in simultaneous interpretation applications for cross-lingual communication in cross-border travel and meetings. Keyword spotting \cite{audhkhasi2017end,rosenberg2017end,kim2019temporal} is employed in short video applications to quickly retrieve relevant content. Additionally, in noisy scenarios, relevant speech tasks \cite{cooke2006audio,harte2015tcd,prajwal2022sub,momeni2020seeing} rely on visual speech to avoid interference from surrounding speech and background noise. 
Despite the growing interest in speech tasks that rely on visual speech, researches \cite{song2022talking,waibel2022face} on visual speech translation are limited and lacks validation due to the lack of multilingual audio-visual speech transcription datasets. This paper proposes a baseline for visual speech translation and introduces the first large-scale audio-visual multilingual translation dataset, AVMuST-TED, which includes 706 hours of audio-visual speech and translation pairs in Spanish, French, Italian, and Portuguese. AVMuST-TED lays a solid foundation or future cross-lingual audio-visual translation tasks, such as Cross-Lingual Talking Head Generation \cite{nvidia2022}.


\subsection{Transfer learning from Audio to Visual}

Many researchers \cite{ren2021learning,shi2022learning,ma2021lira} attempt to enhance the representation of visual speech by leveraging corresponding audio speech, as the two are paired parallel speech streams. Some \cite{ren2021learning,shi2022learning,ma2021lira} use knowledge distillation to bootstrap the training of visual speech models using audio speech models, while others \cite{zhao2020hearing,ma2021lira} have proposed various distillation strategies to optimize the representation of visual speech by mining the intrinsic connection between audio and visual speeches. Some \cite{shi2022learning} also use self-supervised learning, with audio as auxiliary supervision for visual utterances, to obtain fine-grained visual representations. The success of these works demonstrates the critical role of audio speech, which has a higher discrimination compared to visual speech, in training visual speech models.
However, previous works face the modality shift problem during knowledge transfer because they start directly from speeches of two different modalities, audio and visual speeches, with a significant modality gap. In this paper, we propose an cross-modality self-learning framework MixSpeech, that uses synthetic mixed speech to regularize visual speech translation for effective cross-modality knowledge transfer, reducing the gap between the two modalities during knowledge transfer.



\begin{figure*}
    \centering
    \includegraphics[scale=0.58]{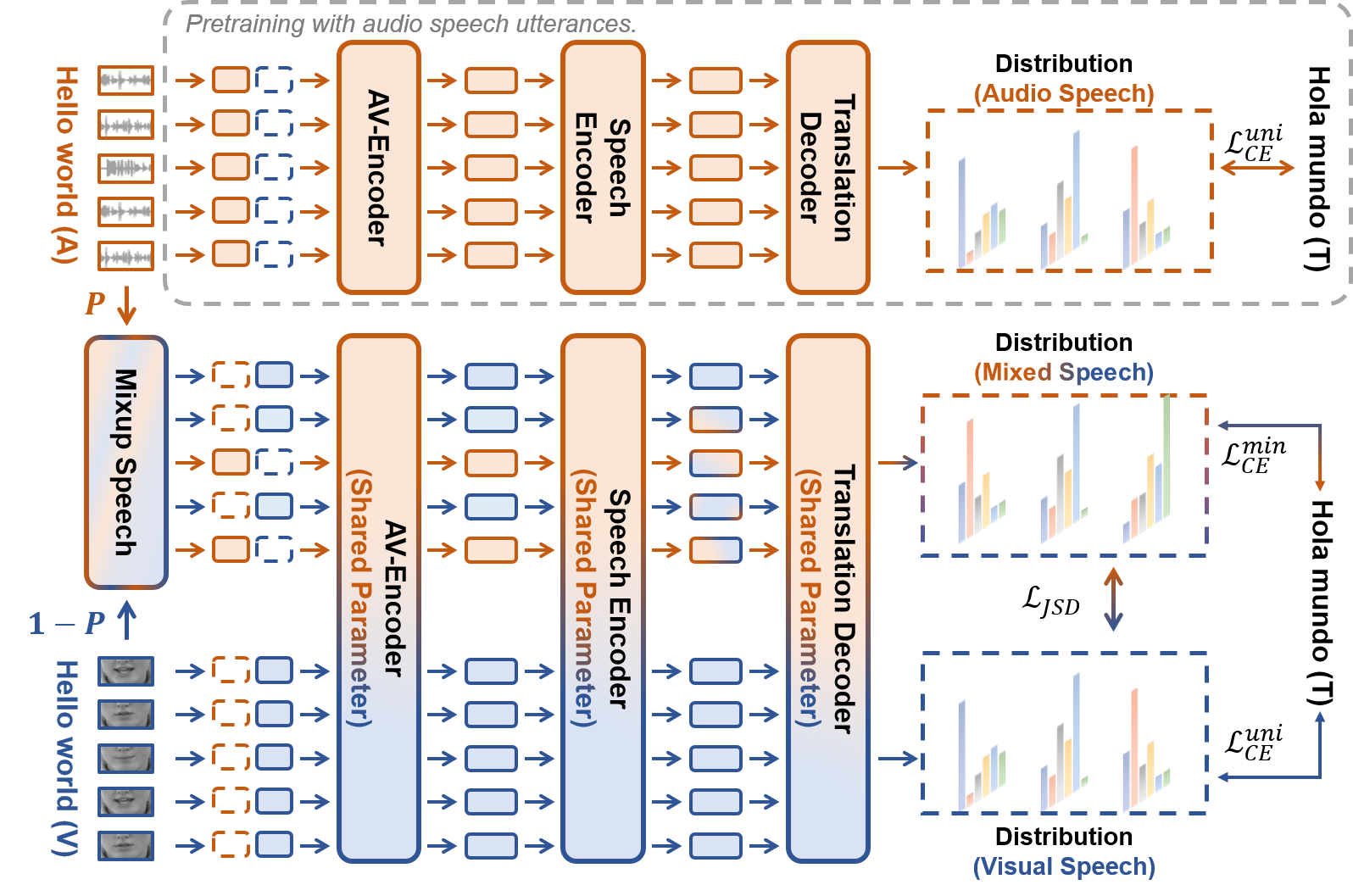}
    \caption{Illustration of our proposed MixSpeech. We first pretrain the model with audio speech translation as shown in the dashed boxed, and then train the visual speech translation with mixed speech regularization. The blank dashed boxes denote the modality missing speech.}
    \label{fig:model}
\end{figure*}

\subsection{Mixup for Cross-Modality Transfer}
Many works \cite{zhang2020does,guo2019mixup,wang2022vlmixer,fang2022stemm,hu2021neural} bridge the gap between modalities with mixup. \cite{zhang2017mixup} proposes mixup for data augmentation to improve model robustness. \cite{chen2020mixtext} suggests mixing at the representation-level to mine implicit associations between labeled and non-labeled sentences. Other works \cite{wang2022vlmixer,tang2021general,fang2022neural,huanggenerspeech} also use mixing to build bridges between different modalities. Some \cite{wang2022vlmixer,fang2022neural} use CLIP \cite{radford2021learning} to retrieve semantically consistent images with text tokens and synthesize mixed sentences for text-visual consistency representation training. Others \cite{tang2021general,fang2022stemm} construct manifold mixup interpolations based on semantic consistency between audio and text to enhance understanding of audio with textual datasets.
By implementing the mixup strategy, these studies have shown notable improvements across a range of tasks, highlighting its potential to facilitate knowledge transfer between different modalities.
However, previous works use fixed hyperparameters \cite{zhang2017mixup} or mapping functions \cite{fang2022stemm} for mixing ratios, which are typically not optimal and cannot be adapted to the training situation.
In this paper, we propose an uncertainty-based \cite{moore1990uncertainty} curriculum learning \cite{bengio2009curriculum} strategy that gradually adjusts mixing ratios and apply mixup strategy for cross-modality knowledge transfer between audio and visual speeches for the first time.

\section{Method}
\subsection{Task Formulation}
As the twin task of speech recognition, speech translation involves translating source language speech into target language text. The speech translation model takes audio speech utterance $\mathbf{A}\texttt{=}{\{\mathbf{A}_t\}}^T_{t=1}\in\mathbb{R}^{T \times D}$ or visual speech utterance $\mathbf{V}\texttt{=}{\{\mathbf{V}_t\}}^T_{t=1}\in\mathbb{R}^{T \times D}$ as input and generates the target language text $\mathbf{w}\texttt{=}{\{\mathbf{w}_i\}}_{i\texttt{=}0}^s$, where $\mathbf{A}_t$ and $\mathbf{V}_t$ represent the $t$-th features in the audio and visual speeches, and $\mathbf{w}_i$ represents the $i$-th word in the target language translation with a total length of $S$. Note that, we stack 4 adjacent acoustic frames together for syncing with visual speech, both with $T$ frames.

\subsection{Overview}
We propose a cross-modal self-learning framework for visual speech translation with audio speech regularization, named MixSpeech, as illustrated in Figure \ref{fig:model}. This model consists of three modules – a feature extractor for extracting speech embeddings, a speech encoder for attending to the contextual dependencies of speech, and a target language-oriented translation decoder. We utilize the pre-trained feature extractor (\texttt{AV-Encoder}) and speech encoder (\texttt{Speech-Encoder}) from the AV-Hubert \cite{shi2022learning} to extract speech representations from both audio and visual speech utterances. Additionally, a randomly initialized translation decoder (\texttt{Trans-Decoder}) is used to autoregressively decode the speech representation into the target language text.
MixSpeech is a two-stage training process: 1) Pretraining the translation decoder with high-discrimination audio speech utterances to learn inter-lingual mapping relations between source language phonemes and target language text, as detailed in subsection \ref{sec:pretrain}. 2) Aligning visual speech with audio speech to transfer the inter-lingual mapping from audio speech to visual in \ref{sec:cross}. The mixed speech \ref{sec:mixed} is synthesized by interpolating audio speech with visual speech in \texttt{MixupSpeech}, bridging the modality gap and enhancing knowledge transfer.



\subsection{Pretraining with Audio Speech}
\label{sec:pretrain}
For uni-modality audio speech $\mathbf{A}\in\mathbb{R}^{T \times D}$ or visual speech $\mathbf{V}\in\mathbb{R}^{T \times D}$, the uni-modality audio-visual feature $\mathbf{e}^{u}\texttt{=}{\{\mathbf{e}^{u}_t\}}^T_{t=1}\in\mathbb{R}^{T \times 2D}$ fed into feature extractor can be defined as:


\begin{equation}
\mathbf{e}^{u}_t=\left\{
\begin{aligned}
\texttt{concat}(\mathbf{0}_{D},\mathbf{V}_t) & , & \mathbf{V}_t  \neq \text{None}, \\
\texttt{concat}(\mathbf{A}_t,\mathbf{0}_{D}) & , & \mathbf{A}_t  \neq \text{None},
\end{aligned}
\right.
\end{equation}
where $\mathbf{0}_{D}$ denotes the feature of missing modality, following the practice of \cite{shi2022learning}. 
And then, we obtain the audio-visual fusion feature $\mathbf{e}^{f}\in\mathbb{R}^{T \times D}$ with \texttt{AV-Encoder}.
The transformer-based \texttt{Speech-Encoder} allows us to obtain the phoneme embedding $\mathbf{e}^p \in \mathbb{R}^{T \times D}$ with the contextual speech details. 
A target language oriented translation decoder \texttt{Trans-Decoder} is appended to autoregressively decode the phoneme embedding $\mathbf{e}^p$ into the target probabilities 
$P^u$,
where $P^u\texttt{=}\{P^u_t\}^S_{t\texttt{=}1}\texttt{=}\{p(\mathbf{w}_t|{\{\mathbf{w}_i\}}^{t-1}_{i=1}, \mathbf{e}^p)\}^S_{t\texttt{=}1}$ represents the probability of the $t$-th word being $\mathbf{w}_t$ when the previous $t-1$ predictions are ${\{\mathbf{w}_i\}}^{t-1}_{i=1}$ and $s$ is the length of the target language translation. During the pretraining, the overall model is trained on audio speech with cross-entropy loss :
\begin{equation}
    \mathcal{L}_{CE}\texttt{=}-\sum_{t=1}^S{\log{p(w_t|{\{w_i\}}^{t-1}_{i=1}, \mathbf{e}^p)}}.
\end{equation}

\subsection{Audio-Visual Speech Mixing}
\label{sec:mixed}
Audio and visual speeches have a huge modality gap, which greatly impacts knowledge transfer across modalities. We attempt to employ mixed speech to bridge two different modalities of speech. 
Since the pair of audio and visual speeches is strictly temporally synchronous, we take advantage of this property to interpolate the mixed speech.
For a pair of synchronized audio and video speech $(\mathbf{A},\mathbf{V})\in\mathbb{R}^{2\times T \times D}$, each visual feature $\mathbf{V}_t$ at $t$-th frame has its corresponding audio feature $\mathbf{A}_t$, representing the same phonetic content.
We interpolate with probability $\phi$ to obtain a mixed speech $\mathbf{e}^{m}\texttt{=}\{\mathbf{e}^{m}_t\}^T_{t=1}\in\mathbb{R}^{T \times 2D}$ derived partly from audio speech and partly from visual speech:

\begin{equation}
\mathbf{e}^{m}_t=\left\{
\begin{aligned}
\texttt{concat}(\mathbf{0}_{D},\mathbf{V}_t) & , & p < \phi, \\
\texttt{concat}(\mathbf{A}_t,\mathbf{0}_{D}) & , & p \geq \phi, 
\end{aligned}
\right.
\end{equation}
where $p$ is sampled from the uniform distribution $U(0,1)$ and $\phi$
is the ratio of speech mixing. 
In particular, we propose a curriculum learning \cite{bengio2009curriculum} based mixing ratio adjustment method that adapts the appropriate $\phi$ as the training progresses. 
The prediction uncertainty \cite{moore1990uncertainty} indicates the confidence of the prediction (smaller is better), and we take it as a signal to adjust the mixing ratio:
\begin{equation}
    \mathbf{u}=\frac{1}{S}\sum_{t=1}^{S} \texttt{Entropy}(P_t).
\end{equation}

If the discrimination of mixed speech is insufficient to regularize visual speech translation and maintain $n$ steps ($\Delta \mathbf{u}\texttt{=}\mathbf{u}^{v}\texttt{-}\mathbf{u}^{m}\texttt{<}k \mathbf{u}^{v}$, where $\mathbf{u}^{v}$ and $\mathbf{u}^{m}$ represent the uncertainty of uni-modality (visual) and mixed speech, respectively, and the threshold hyperparameter $k$ is set to 0.05 with $n$ set to 20 in our work), we gradually increase the proportion of audio at a rate of $\alpha$ ($\phi'\texttt{=}\alpha\,\phi$). We initialize $\phi\texttt{=}0.1$ to prevent excessive initial modality gap and maintain $\phi \in \left[0.1,0.9\right]$ throughout the training process.






\subsection{Cross-Modality Self-Learning for Speech}
\label{sec:cross}
Since audio speech is more distinguished compared to visual speech, we intend to boost visual speech translation with the knowledge from audio speech. And the mixed speech bridges the gap between audio speech and visual speech, allowing us to boost cross-modality knowledge transfer with it. With audio speech feature $\mathbf{A}\in\mathbb{R}^{T \times D}$ and visual speech feature $\mathbf{V}\in\mathbb{R}^{T \times D}$ fed into the modules with shared parameters, the uni-modality visual speech feature $\mathbf{e}^{u}$ and the mixed speech feature $\mathbf{e}^{m}$ are decoded into the target probabilities $P^{u}$ and $P^{m}$, respectively. 

After the pre-training with audio speech translation, the model is promising enough for mixed speech containing partial audio speech, we adopt the Jensen-Shannon Divergence (JSD) \cite{menendez1997jensen} to regularize the probabilities of these two different speeches:
\begin{equation}
    \mathcal{L}_{JSD}=\sum_{t=1}^S JSD(P_t^{m}\Vert P_t^{u}).
\end{equation}

As this probability is across the entire training vocabulary, we are able to perform fine-grained regularization to enhance the training of visual speech. Meanwhile, we also minimize the cross-entropy loss between two speech translations and the real translation, $\mathcal{L}\texttt{=}\mathcal{L}_{CE}^{uni}\texttt{+}\lambda_1\mathcal{L}_{CE}^{mix}\texttt{+}\lambda_2\mathcal{L}_{JSD}$,
where $\lambda_1$ and $\lambda_2$ are hyperparameters of loss weights, while $\lambda_1\texttt{=}\lambda_2\texttt{=}1.0$ in this work.

\begin{table*}[tb]
\begin{center}
\begin{tabular}{@{}rcccccrrrrr@{}}
\toprule
\multicolumn{1}{c}{}                                   & \multicolumn{5}{c}{\textbf{Target Language Hours}}                  & \multicolumn{1}{c}{}                                  & \multicolumn{1}{c}{}                                                     & \multicolumn{1}{c}{}                                                       & \multicolumn{2}{c}{\textbf{\# $\sum{\textbf{Tokens}}$}}             \\ \cmidrule(lr){2-6} \cmidrule(l){10-11} 
\multicolumn{1}{c}{\multirow{-2}{*}{\textbf{Dataset}}} & \textbf{En} & \textbf{Es} & \textbf{Fr} & \textbf{It} & \textbf{Pt} & \multicolumn{1}{c}{\multirow{-2}{*}{\textbf{\# Lang}}} & \multicolumn{1}{c}{\multirow{-2}{*}{\textbf{\#   $\sum{\textbf{Hrs}}$}}} & \multicolumn{1}{c}{\multirow{-2}{*}{\textbf{\#   $\sum{\textbf{Sents}}$}}} & \multicolumn{1}{c}{\textbf{src}} & \multicolumn{1}{c}{\textbf{tgt}} \\ \midrule
\multicolumn{11}{l}{\cellcolor[HTML]{C0C0C0}Audio-Only}                                                                                                                                                                                                                                                                                                                                                            \\ 
LibriSpeech   \cite{panayotov2015librispeech}          & 960h        & -           & -           & -           & -           & 1                                                     & 960h                                                                     & 180K                                                                           & 5.9M                                 & 5.9M                                 \\
MuST-C \cite{di2019must}                               & -           & 504h        & 492h        & 465h        & 385h          & 8                                                     & 3\,617h                                                                    & 2\,016K                                                                      & 38.1M                            & 35.8M                            \\
VoxPopuli   \cite{wang2021voxpopuli}                   & 543h        & 441h        & 427h        & 461h        &-        & 16                                                    & 5\,967h                                                                    & 2\,045K                                                                      & 65.0M                            & 60.1M                            \\ \midrule
\multicolumn{11}{l}{\cellcolor[HTML]{C0C0C0}Audio-Visual}                                                                                                                                                                                                                                                                                                                                                          \\ 
LRS2   \cite{afouras2018deep}                          & 224h        & -           & -           & -           & -           & 1                                                     & 224h                                                                     & 143K                                                                       & 2.3M                             & 2.3M                             \\
LRS3 \cite{afouras2018lrs3}                            & 433h        & -           & -           & -           & -           & 1                                                     & 433h                                                                     & 151K                                                                       & 4.2M                             & 4.2M                             \\
AVMuST-TED (ours)                                       & -           & 198h        & 185h        & 165h        & 158h        & 4                                                     & 706h                                                                     & 925K                                                                       & 7.3M                             & 7.0M         \\ \bottomrule
\end{tabular}

\end{center}
\caption{Comparison of audio-visual speech recognition/translation datasets. \#Lang denotes the number of target languages. \#$\sum{\textbf{Hrs}}$ denotes the overall duration of speech in the dataset, \#Sents and \#Tokens denote the overall sentences and the overall token, respectively.}
\label{tab:dataset}
\end{table*}
\section{Experiments}

\subsection{Datasets}
\textbf{AVMuST-TED}.
To obtain a corpus for AVST, we screened a set of TED and TEDx talks with multilingual subtitles as the data source. All transcriptions and translations are performed strictly following the TED Translation Guidelines and require collaboration between at least one translator (or transcriber) and one reviewer. The prior lip-reading dataset acquisition pipeline is followed to crop face-tracks, and an audio-visual alignment network, SyncNet, is adopted for speaker proofreading. Table \ref{tab:dataset} compares AVMuST-TED with related datasets, and it is the first audio-visual speech translation dataset containing translations from English (En) to four target languages: Spanish (Es), French (Fr), Italian (It), and Portuguese (Pt). These four languages have the most translated subtitles in TED, and 1024/1536 pieces of data are randomly sampled for each language as the \textit{test}/\textit{validation} set. The information about AVMuST-TED is detailed in Appendix \ref{sec:avmust}.

\textbf{LRS2\&3} \cite{afouras2018deep,afouras2018lrs3}, two commonly used publicly available English wild audio-visual speech recognition datasets, are adopted to demonstrate the lip-reading performance, containing 224 hours of video from BBC television shows and 433 hours of video from TED and TEDx talks. The training data in both datasets is divided into two partitions, namely \textit{Pretrain} and \textit{Train}, both of which are transcribed from videos to text at the sentence level. The only difference is that the video clips in the \textit{Pretrain} partition are not strictly trimmed and sometimes longer than the corresponding text. In our experiments, we employ different amounts of training data from LRS2 and LRS3, including \textit{Pretrain+Train} (224/433h) for high resource and \textit{Train} (29/30h) for low resource.


 \textbf{CMLR} \cite{zhao2019cascade}, widely used dataset for Mandarin audio-visual speech recognition, contains 61 hours audio-visual speech utterances collected from Chinese TV stations. In our experiments, we adopt this dataset to demonstrate the performance of our proposed MixSpeech in low-resource languages such as Mandarin. Additionally, we sample a training set containing only 12 hours of utterances in the manner of \cite{zhao2020hearing} for low resource scenario.


\subsection{Evaluation and Implementation Details}

In this paper, we measure the performance of MixSpeech on two speech tasks, speech recognition and speech translation.
For speech recognition, word error rate (WER) is adopted as the evaluation metric, which is defined as $\texttt{WER}\textit{=}{(S+D+I)}/{M}$, where \(S,D,I,M\) represent the number of words replaced, deleted, inserted, and referenced.
As for speech translation, the case-sensitive detokenized BLEU score is computed using \textsc{Sacre}BLEU \cite{post-2018-call}, following the same evaluation methodology as in previous speech translation works \cite{di2019must,wang2021voxpopuli}. The implementation details are provided in Appendix \ref{app:implement} due to page limitations.


\begin{table}[tb]
\tabcolsep=2.5pt
\begin{center}
\begin{tabular}{@{}llcccc@{}}
\toprule
\multicolumn{1}{c}{\multirow{2}{*}{\textbf{Method}}} & \multirow{2}{*}{\textbf{M}} & \multicolumn{4}{c}{\textbf{BLEU $\uparrow$}}                    \\ \cmidrule(l){3-6} 
\multicolumn{1}{c}{}                                 &                             & \textbf{En-Es} & \textbf{En-Fr} & \textbf{En-It} & \textbf{En-Pt} \\ \midrule
Cascaded                                             & V                           & 12.7           & 11.3           &  11.5          &  13.2              \\
AV-Hubert \cite{shi2022learning}                     & V                           & 14.2           & 12.6           &  12.9          &  14.8              \\
Cascaded                                             & A$_{\texttt{(+Noise)}}$     & 16.0           & 12.9           &  12.6          &  15.5              \\
AV-Hubert \cite{shi2022learning}                     & A$_{\texttt{(+Noise)}}$     & 17.6           & 14.5           &  14.1          &  17.1              \\
MixSpeech(ours)                                      & V                           & \textbf{18.5}  & \textbf{15.1}  & \textbf{14.3}               & \textbf{17.2}               \\ \bottomrule
\end{tabular}
\end{center}
\caption{Comparison of the performances of visual speech translation on AVMuST-TED with those of the noisy audio speech translation. The results of noisy audio speech translation are the mean value at five SNRs \{-20, -10, 0, 10, 20\}db.}
\label{tab:vst}
\end{table}

\subsection{Performance of Speech Translation}
\textbf{End-To-End Models VS. Cascaded Models.} Table \ref{tab:vst} presents a comparison of the lip translation performance between two representative methods: 1) an end-to-end model, implemented based on the state-of-the-art AV-Hubert \cite{shi2022learning} method for visual speech-related tasks, and 2) a cascaded model, combining a speech recognition model (\ie, Lip-Reading or ASR) with a machine translation model. In the cascaded model, we use the speech recognition model trained by AV-Hubert on LRS3, which achieve the best lip-reading performance to date, and a transformer-based machine translation model trained on the paired translated text corpus in AVMuST-TED.
Comparing the lip translation performance of the end-to-end model and the cascade model, we find that the BLEU score of the end-to-end model improved by +1.3 to +1.6. This result demonstrates that the end-to-end trained model can effectively prevent the accumulation of errors caused by the model cascade, and that lip translation cannot be simply disassembled as the superposition of lip reading and machine translation.

\textbf{MixSpeech VS. Prior Methods.}
Due to the discrimination of speech between modalities, visual speech models are not able to translate speech content as accurately as audio speech models. 
To address the issue of low discrimination in visual speech, we propose MixSpeech, which is a cross-modality self-learning framework that employs mixed speech to transfer knowledge obtained from audio speech pre-training into the visual speech model. Our proposed MixSpeech significantly improves the BLEU score by another +1.4 to +4.3.
Furthermore, the improvement from MixSpeech is related to the discrepancy in speech translation between audio and visual modalities. For example, En-Es exhibits a larger discrepancy of 14.7 between audio and visual speech translation, ranging from 28.9 to 14.2, and MixSpeech significantly improves it by +4.3. Conversely, Italian shows a smaller discrepancy of 10.9, ranging from 23.8 to 12.9, and improves only by +1.4. This highlights that the improvement in lip translation stems from the knowledge acquired from audio speech translation.

\textbf{Visual Speech VS. Noisy Audio Speech.}
We also evaluate the performance of audio speech translation in noisy environments, by adding noise sampled from MUSAN \cite{Snyder2015MUSANAM} to the audio speech and measuring the performance at five SNR levels \{-20,\,-10,\,0,\,10,\,20\}db. We compare the average BLEU scores of different SNRs and present the detailed performance in Appendix \ref{app:noisy_speech}. Our experiments show that although noisy audio speech performs better than visual speech, the translation performance is still significantly lower compared to noiseless audio speech. In contrast, MixSpeech, which fully leverages the knowledge of audio speech, greatly improves the visual speech translation performance, making it more reliable in noisy scenes. We also provide a comparison of translation with audio speech and audio-visual speech, demonstrating that visual speech enhances the ceiling and robustness of speech translation, but the details are only available in the Appendix \ref{app:noisy_speech} since audio-visual speech does not require the cross-modality knowledge transfer proposed in this paper.

\begin{table}[tb]
\tabcolsep=4pt
\begin{center}
\begin{tabular}{@{}clccc@{}}
\toprule
\multirow{2}{*}{\textbf{\# RES}} & \multicolumn{1}{c}{\multirow{2}{*}{\textbf{Method}}} & \multicolumn{3}{c}{\textbf{$\textsc{WER}_{\textit{(Labeled Visual Utts Hrs)}}\downarrow$}}                                     \\ \cmidrule(l){3-5} 
                                    & \multicolumn{1}{c}{}                                 & \textbf{CMLR}                   & \textbf{LRS2}                    & \textbf{LRS3}                   \\ \midrule
\multirow{11}{*}{\textbf{High}}     & \small{WAS}   \cite{son2017lip}                      & $\textsc{38.9}_{(61)}$          & $\textsc{70.4}_{(224)}$          & -                               \\
                                    & \small{TM-seq2seq} \cite{afouras2018deep}            & -                               & $\textsc{49.8}_{(698)}$          & $\textsc{59.9}_{(698)}$         \\
                                    & \small{CSSMCM} \cite{zhao2019cascade}                & $\textsc{32.5}_{(61)}$          & -                                & -                               \\
                                    & \small{Conv-seq2seq} \cite{zhang2019spatio}          & -                               & $\textsc{51.7}_{(698)}$          & $\textsc{60.1}_{(698)}$         \\
                                    & \small{CTC+KD} \cite{afouras2020asr}                 & -                               & $\textsc{51.3}_{(224)}$          & $\textsc{58.9}_{(433)}$         \\
                                    & \small{LIBS} \cite{zhao2020hearing}                  & $\textsc{31.3}_{(61)}$          & $\textsc{65.3}_{(698)}$          & -                               \\
                                    & \small{CTCH} \cite{ma2020transformer}                & $\textsc{22.0}_{(61)}$          & -                                & -                               \\
                                    & \small{Master} \cite{ren2021learning}                & -                               & $\textsc{49.2}_{(698)}$          & $\textsc{59.0}_{(698)}$         \\
                                    & \small{Sub-Word} \cite{prajwal2022sub}               & -                               & $\textsc{28.9}_{(698)}$          & $\textsc{40.6}_{(698)}$         \\
                                    & $\dagger$\small{AV-Hubert} \cite{shi2022learning}             & $\textsc{12.7}_{(61)}$          & $\textsc{28.7}_{(224)}$          & $\textsc{28.6}_{(433)}$         \\
                                    & \small{MixSpeech(ours)}                                         & \textbf{$\textsc{11.1}_{(61)}$} & \textbf{$\textsc{25.5}_{(224)}$} & \textbf{$\textsc{28.0}_{(433)}$}   \\ \midrule
\multirow{3}{*}{\textbf{Low}}       & \small{LIBS}   \cite{zhao2020hearing}                & $\textsc{50.5}_{\textcolor{blue}{(12)}}$          & -                                & -                               \\
                                    & $\dagger$\small{AV-Hubert} \cite{shi2022learning}             & $\textsc{25.8}_{\textcolor{blue}{(12)}}$          & $\textsc{31.4}_{\textcolor{blue}{(29)}}$           & $\textsc{32.5}_{\textcolor{blue}{(30)}}$          \\
                                    & \small{MixSpeech(ours)}                                         & \textbf{$\textsc{18.5}_{\textcolor{blue}{(12)}}$} & \textbf{$\textsc{26.9}_{\textcolor{blue}{(29)}}$}  & \textbf{$\textsc{28.6}_{\textcolor{blue}{(30)}}$} \\ \bottomrule
\end{tabular}
\end{center}
\caption{Comparison of lip reading methods under different resource conditions. \# RES represents the amount of resources. \textcolor{blue}{(Hours)} highlighted in blue are used for low resources. $\dagger$ For better comparison, we reproduce AV-Hubert on CMLR and LRS2.}
\label{tab:vsr}
\end{table}

\subsection{Performance of Speech Recognition}

As shown in Table \ref{tab:vsr}, we compare the performance of MixSpeech on another visual speech task, lip reading (\ie, Visual Speech Recognition), to highlight the mixspeech from more perspectives.
MixSpeech obtain state-of-the-art performance on three datasets, two for English (25.5\% on LRS2 and 28.0\% on LRS3) and one for Chinese (11.1\% on CMLR), demonstrating that this cross-modality self-learning framework can be applied for different languages to capture the intrinsic association between audio and visual speeches and thus effectively improve the understanding of visual speech.
Since visual speech is relatively low-resource, we verify whether MixSpeech can effectively improve the performance of visual speech tasks in low-resource with audio speech. Compared with previous methods, MixSpeech boosts the WER of lip-reading by -3.9\% to -7.3\%, highlighting the critical role of high-resource audio speech in low-resource visual speech tasks. Specifically, on LRS2 and LRS3, the performance of Mixspeech in the low-resource scenario (26.9\%/28.6\% WER obtained with only 29h/30h visual utterances) outperforms the performances of prior methods in the high-resource scenario (28.7\%/28.6\% obtained with 224h/433h or even more visual utterances). Even though with only limited labeled visual corpus, our proposed MixSpeech performs no less than works with more. It is the bridge between two modalities of speech, which helps visual speech to access the knowledge stored in high-resource and high-discrimination audio speech without barriers.



\begin{figure}[b]
    \centering
    \includegraphics[scale=0.58]{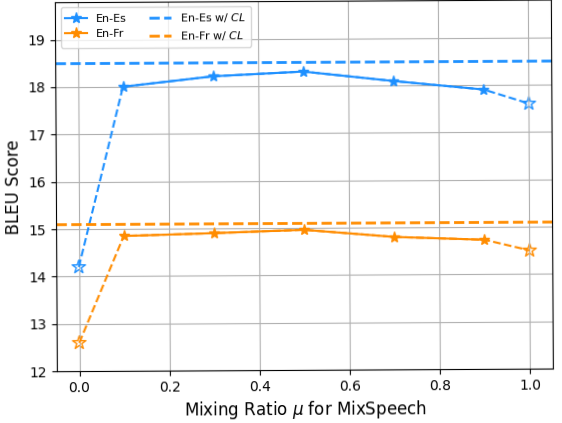}
    \caption{BLEU scores of MixSpeech with different speech regularization on En-Es and En-Fr. $\phi$ = 0: no audio speech regularization, $\phi \in \left(0,1\right)$: mixed speech regularization, $\phi$ = 1: only audio speech regularization. The dashed lines represent the adjustable mixing ratio strategy based on curriculum learning.}
    \label{fig:mix_ratio}
\end{figure}

\subsection{Can MixSpeech bridge cross-modality speech?}
Our proposed MixSpeech builds a bridge between cross-modality speech through cross-modality self-learning, with the properly mixes speech. The details are as follows:

\textbf{Cross-Modality Self-Learning for Knowledge Transfer.}
The experiments in Figure \ref{fig:mix_ratio} provide a positive answer to the question of whether MixSpeech can contribute to achieving knowledge transfer between audio and visual speeches. We evaluate the performance of visual speech translation with different regularization strategies: no audio speech regularization (\ie, $\phi$ = 0), mixed speech regularization with different mixing ratios (\ie, $\phi \in \left(0,1\right)$, audio speech regularization (\ie, $\phi$ = 1), and mixing ratio adjustable mixed speech regularization (\ie, dashed lines).
It is evident that the cross-modality self-learning framework significantly enhances visual speech translation, as all performances with audio speech regularization are noticeably better than those without self-learning ($\phi$ = 0), demonstrating the effectiveness of our proposed MixSpeech.

\textbf{Narrow the Cross-Modality Distance with Properly Mixed Speech.}
Moreover, the introduction of mixed speech facilitates smoother cross-modality knowledge transfer by narrowing the modality gap between speeches. Some segments in the mixed speech come from the visual speech, making it much closer to visual speech in terms of modality distance than audio speech.
When regularizing with mixed speech in En-Es, the translation performance of visual speech improves further by +0.3 to +0.8 compared to audio speech regularization alone. Among them, bootstrapping with mixed speech of mixing ratio $\phi$ = 0.5 achieves the highest BLEU score of 18.3. This demonstrates that a reasonably mixed ratio ensures that it is neither overly biased towards visual speech, leading to a lack of knowledge of audio speech, nor overly biased towards audio speech, leading to excessive cross-modality distances that affect knowledge transfer.
The adjustable mixing ratio strategy based on curriculum learning further increases the applicability of mixed speech to cross-modality self-learning training, thereby boosting visual speech translation performance again.

\begin{table}[h]
\tabcolsep=3pt
\begin{center}
\begin{tabular}{@{}c|ccc|cccc@{}}
\toprule
\multirow{2}{*}{\textbf{ID}} & \multicolumn{3}{c|}{\textbf{Method}} & \multicolumn{4}{c}{\textbf{BLEU $\uparrow$}} \\ \cmidrule(l){2-8} 
                     &$\mathcal{L}_{CE}^{mon}$ & $\mathcal{L}_{CE}^{mix}$ & $\mathcal{L}_{JSD}$ & \textbf{En-Es} & \textbf{En-Fr} & \textbf{En-It} & \textbf{En-Pt}    \\ \midrule
\#1                  & \ding{52}    &               &                             & 14.2    & 12.6    & 12.9    & 14.8   \\
\#2                  & \ding{52}    & \ding{52}     &                             & 17.5    & 14.3    &  13.6   & 16.5   \\
\#3                  & \ding{52}    &               & \ding{52}                   & 18.1    & 14.8    &  14.1   & 16.9   \\
\#4                  & \ding{52}    & \ding{52}     & \ding{52}         & \textbf{18.5}    & \textbf{15.1}    & \textbf{14.3}    & \textbf{17.2}  \\ 
\bottomrule
\end{tabular}
\end{center}
\caption{BLEU of different module combinations in MixSpeech.}
\label{tab:loss}
\end{table}

\subsection{What role does each part play in MixSpeech?}
The effectiveness of MixSpeech, which is a cross-modality self-learning framework designed to improve visual translation performance, has been demonstrated. In this study, we investigate the role of each component in detail and present relevant experiments in Table \ref{tab:loss}:

\textbf{Bridging the cross-modality gaps.} We observe a significant improvement in the lip translation performance with the inclusion of $\mathcal{L}_{JSD}$ (ID: \#3, \#4) for regularizing the probabilities of visual speech and mixed speech, compared to without (ID: \#1, \#2). 
Specifically, experiment \#3 with $\mathcal{L}_{JSD}$ outperform experiment \#2 with $\mathcal{L}_{CE}^{mix}$ by +0.6 in lip translation performance on En-Es. This demonstrates that $\mathcal{L}_{JSD}$ is the main contributor to achieving cross-modality knowledge transfer by building a bridge between the two speeches and performing fine-grained regularization across the probability of each word.

\textbf{Maintaining knowledge of audio speech .} It is also important to note that during the regularization process, the representation of audio speech is also affected by visual speech, which can interfere with the knowledge of audio speech and ultimately harm the lip translation performance of MixSpeech. As evidenced by experiment \#2, the lip translation performance on En-Es decrease by -0.4 compared to experiment \#3 when $\mathcal{L}_{CE}^{mix}$ is not applied. To address this issue, $\mathcal{L}_{CE}^{mix}$ is introduced to enhance the training ceiling of the cross-modality self-learning framework. By maintaining the translation performance of mixed speech and preventing the excessive disturbance to audio speech knowledge, $\mathcal{L}_{CE}^{mix}$ helps to improve the overall performance of MixSpeech.


\begin{table*}[tb]
\tabcolsep=2pt
\renewcommand{\arraystretch}{0.7}
\begin{center}
\begin{tabular}{@{}l|crl@{}}
\toprule[1.5pt]
\multirow{5}{*}{\textbf{En-Es}} & \multicolumn{3}{r}{\begin{minipage}[b]{0.93\linewidth}\includegraphics[scale=0.75]{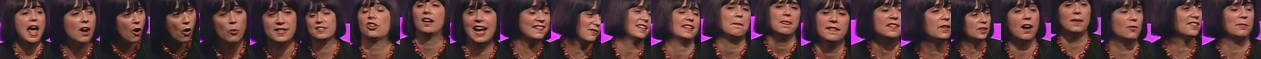}\end{minipage}} \\ \cmidrule(l){2-4} 
                       & \multirow{2}{*}{\textbf{En}}                     & \small{Transcription:}       & and always as a child I had this fantasy that somebody would come and rescue me    \\
                       &                                 & \small{MixSpeech:}       & and always as a child I had this fantasy that somebody would come and rescue \textcolor{lightgray}{me}    \\\cmidrule(l){2-4} 
                       & \multirow{2}{*}{\textbf{Es}}    & \small{Ground Truth:}       &  y de niña siempre tenía la fantasía de que alguien vendría a salvar me de    \\
                       &                                 & \small{MixSpeech:}     &  y \textcolor{lightgray}{de} \textcolor{blue}{(desde)} \textcolor{lightgray}{niña} \textcolor{red}{\sout{niño}} siempre tenía \textcolor{lightgray}{la} \textcolor{blue}{esta} fantasía de que alguien vendría \textcolor{lightgray}{a salvar} me \textcolor{blue}{(rescató)} de   \\ 
\midrule[1.pt]
\multirow{5}{*}{\textbf{En-Fr}} & \multicolumn{3}{r}{\begin{minipage}[b]{0.93\linewidth}\centering\includegraphics[scale=0.75]{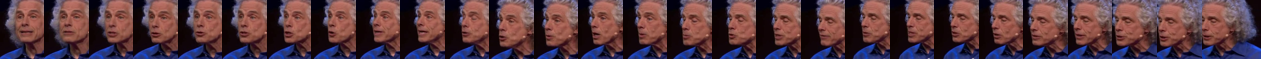}\end{minipage}} \\ \cmidrule(l){2-4} 
                       & \multirow{2}{*}{\textbf{En}}                     & \small{Transcription:}       & and solutions create new problems which have to be solved in their turn    \\
                       &                                 & \small{MixSpeech:}       & \textcolor{lightgray}{and} solutions to create new problems which have to be solved in \textcolor{lightgray}{their turn} \textcolor{red}{\sout{this year}}    \\ \cmidrule(l){2-4} 
                       & \multirow{2}{*}{\textbf{Fr}}    & \small{Ground Truth:}       &  et les solutions créent de nouveaux problèmes devant être résolus à leur tour   \\
                       &                                 & \small{MixSpeech:}     &     et \textcolor{lightgray}{les} \textcolor{blue}{(des)} solutions \textcolor{red}{\sout{pour}} créer de nouveaux problèmes \textcolor{lightgray}{devant} \textcolor{red}{\sout{qui doivent}} être résolus à leur tour\\ 
\midrule[1.pt]
\multirow{5}{*}{\textbf{En-It}} & \multicolumn{3}{r}{\begin{minipage}[b]{0.93\linewidth}\includegraphics[scale=0.75]{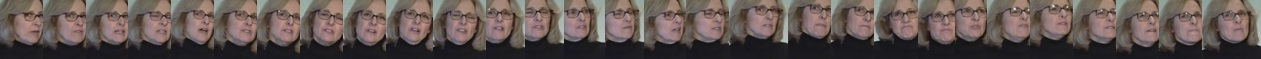}\end{minipage}} \\ \cmidrule(l){2-4} 
                       & \multirow{2}{*}{\textbf{En}}                     & \small{Transcription:}       & one of the last 10,000 years, and the other certainly of the last 25 years    \\
                       &                                 & \small{MixSpeech:}       & one of the last 10,000 years\textcolor{lightgray}{,} and the other \textcolor{lightgray}{certainly} \textcolor{red}{\sout{assistance}} of the last 25 years    \\ \cmidrule(l){2-4} 
                       & \multirow{2}{*}{\textbf{It}}    & \small{Ground Truth:}       & uno presente negli ultimi diecimila anni e l’altro certamente negli ultimi 25 anni    \\
                       &                                 & \small{MixSpeech:}     & uno presente negli ultimi \textcolor{lightgray}{diecimila} \textcolor{blue}{(10000)} anni e l'altro certamente negli ultimi 25 anni    \\ 
\midrule[1.pt]
\multirow{4}{*}{\textbf{En-Pt}} & \multicolumn{3}{r}{\begin{minipage}[b]{0.93\linewidth}\includegraphics[scale=0.75]{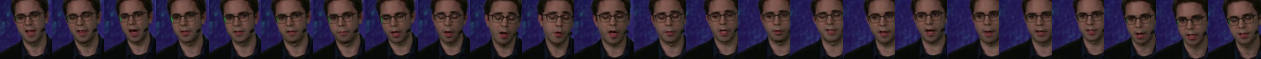}\end{minipage}} \\ \cmidrule(l){2-4} 
                       & \multirow{2}{*}{\textbf{En}}                     & \small{Transcription:}       & has around 2,000 people descending on MIT's campus    \\
                       &                                 & \small{MixSpeech:}       & \textcolor{lightgray}{has} around 2,000 people descending on MIT's campus        \\ \cmidrule(l){2-4} 
                       & \multirow{2}{*}{\textbf{Pt}}    & \small{Ground Truth:}      & congregam cerca de 2000 pessoas no campus do mit    \\
                       &                                 & \small{MixSpeech:}     & \textcolor{lightgray}{congregam} \textcolor{red}{\sout{de}} cerca de 2000 pessoas \textcolor{red}{\sout{a estudar}} no campus do mit    \\ 
\bottomrule[1.5pt]
\end{tabular}
\end{center}
\caption{Qualitative performance of Visual Speech Recognition and Translation on AVMuST-TED. \textcolor{red}{\sout{Red Strikeout Words}}: mistranslated words with opposite meaning, \textcolor{blue}{(Blue Words in parentheses)}: mistranslated words with similar meaning, \textcolor{gray}{Gray Words}: the absent words.}
\label{tab:quan}
\end{table*}

\subsection{Qualitative results}
We present several examples of lip translation in Table \ref{tab:quan} to qualitatively evaluate the translation quality of MixSpeech. The translation results are very close to the ground truth, and the semantics are consistent. We observe two types of words that differ in translation: synonyms and context-sensitive translations. Synonyms that have different spellings but the same meaning, such as \texttt{salvar} and \texttt{rescató} in Spanish, both meaning `rescue', and \texttt{diecimila} and \texttt{10000} in Italian, both meaning `ten thousand', are commonly found in translation tasks and can affect translation consistency. Additionally, there are translations that require context information, such as when the speaker refers to themselves as a \texttt{child}, and the translation in Spanish needs to take into account the speaker's gender to choose between \texttt{niña} for girl' or \texttt{niño} for boy' and `child'. The qualitative translation results of MixSpeech demonstrate its capability to achieve reliable cross-lingua lip translation. In Appendix \ref{app:qua}, we also provide translation results of noisy audio speech translation with visual speech translation and audio speech translation with audio-visual speech translation to highlight the importance of visual speech in speech translation.


\section{Conclusion}
With the advancement of online technologies, such as online healthcare and sales, language barriers often prevent these tools from reaching and benefiting disadvantaged areas. In light of this, we focus on visual speech, a branch of the speech stream, and aim to translate visual speech from source languages to other target languages for cross-linguistic communication, specifically through lip translation. We meticulously curate the AVMuST-TED dataset, consisting of 706 hours of speech clips with professional translations from TED, to facilitate cross-linguistic research on visual speech. 
We also introduce MixSpeech, a cross-modality self-learning framework that utilizes mixed speech to regularize visual speech translation and achieves state-of-the-art performance in lip translation on AVMuST-TED and lip reading on LRS2, LRS3, and CMLR datasets.

Moreover, our work on visual speech and AVMuST-TED lay a solid foundation for further research on visual speech in cross-lingual fields. There are numerous related tasks with great potential for practical applications, such as Cross-Lingual Talking Head Generation \cite{nvidia2022}. These tasks hold immense promise for breaking down language barriers and promoting communication across diverse communities.



{\small
\bibliographystyle{ieee_fullname}
\bibliography{main}
}

\newpage
\,
\newpage
\appendix

\section{AVMuST-TED}
\label{sec:avmust}

\subsection{Details of AVMuST-TED}
The dataset consists of over 706 hours of video, extracted from 4598 TED and TEDx talks in English. The visual speech corpus is provided as face-centered video in \texttt{.avi} files with a resolution of $224\times224$ and a frame rate of 25fps. The audio speech corpus is provided as the single-track, 16-bit 16kHz \texttt{.wav} files. Each pair of audio and video speech has its corresponding translation into other languages. Following the previous workflow \cite{afouras2018deep,yang2019lrw,afouras2018lrs3} of visual-speech dataset acquisition, we fetch the complete face track from the massive data \cite{lienhart2001reliable} and perform audio-visual synchronization testing to determine whether it is the face track of the speaker \cite{chung2017out}. We take the four most amount of translation pairs, En-Es, En-Fr, En-It and En-Pt, from the numerous translation combinations of TED, and the detailed statistics in four different languages at AVMuST-TED are shown in Table \ref{tab:detail_AVMustted}.

\begin{table}[h]
\begin{center}
    \begin{tabular}{@{}lcccc@{}}
    \toprule
    \textbf{Target Language} & \textbf{Hours} & \textbf{Sents} & \textbf{Vocab} & \textbf{Tokens} \\ \midrule
    Spanish (Es)    & 198h  & 258K  & 95K   & 2.0M   \\
    French (Fr)     & 185h  & 244K  & 91K   & 1.9M   \\
    Italian (It)    & 165h  & 218K  & 95K   & 1.6M   \\
    Portuguese (Pt) & 158h  & 205K  & 84K   & 1.5M   \\ \bottomrule
    \end{tabular}
\end{center}
\caption{Statistics in four different languages at AVMuST-TED.}
\label{tab:detail_AVMustted}
\end{table}

\begin{figure}[b]
    \begin{center}
        \includegraphics[scale=0.22]{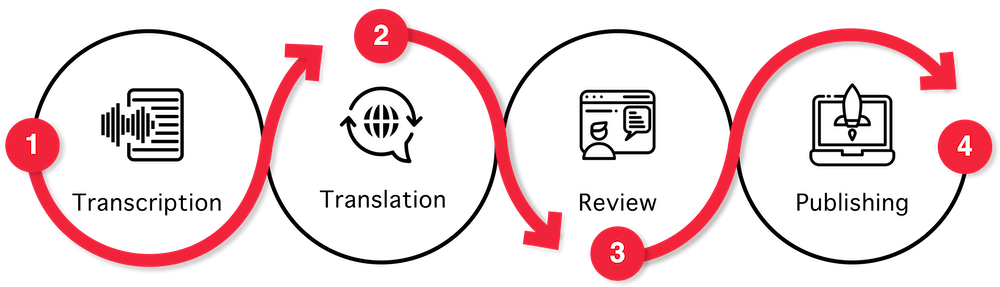}
    \end{center}
    \caption{The TED translation workflow before publication.}
    \label{fig:ted_trans}
\end{figure}

\begin{table}[tb]
\begin{center}
\begin{tabular}{@{}ll@{}}
\toprule[1.2pt]
\textbf{En} & \small{It's a shared database} \\ \midrule
\textbf{Es} & \small{Es una base de datos compartida} \\
\textbf{Fr} & \small{C'est une base de données partagée} \\
\textbf{It} & \small{È una base dati condivisa} \\
\textbf{Pt} & \small{É uma base de dados partilhada} \\
\midrule[1.2pt]
\textbf{En} & \small{That object was about 10 kilometers across} \\ \midrule
\textbf{Es} & \small{Ese objeto tenía un diámetro de 10 km} \\
\textbf{Fr} & \small{Cet objet mesurait dix kilomètres de largeur environ} \\
\textbf{It} & \small{Quell'oggetto aveva un diametro di circa 10 chilometri} \\
\textbf{Pt} & \small{Esse objeto tinha cerca  de 10 km de diâmetro} \\ 
\midrule[1.2pt]
\textbf{En} & \small{Can I correct my boss when they make a mistake?}\\ \midrule
\textbf{Es} & \small{¿Puedo corregir a mi jefe cuando comete un error?} \\ 
\textbf{Fr} & \small{Puis-je corriger mon patron quand il fait une erreur ?} \\
\textbf{It} & \small{Posso correggere il mio capo quando fa un errore?} \\
\textbf{Pt} & \small{Posso corrigir o meu chefe quando ele comete um erro?} \\ 
\midrule[1.2pt]
\textbf{En} & \small{Now this turns out to be surprisingly common} \\ \midrule
\textbf{Es} & \small{Ahora bien, esto resulta ser  sorprendentemente común} \\
\textbf{Fr} & \small{Il s'avère que cela soit surprenamment commun} \\
\textbf{It} & \small{Ora questo risulta essere sorprendentemente comune} \\
\textbf{Pt} & \small{Isto parece ser surpreendentemente comum} \\
\bottomrule[1.2pt]
\end{tabular}

\end{center}
\caption{Examples of the source language transcription (En) and target language translation (Es, Fr, It, Pt) for audio-visual speeches (En) in AVMuST-TED.}
\label{tab:translation_text}
\end{table}

\begin{table*}[t]
\begin{center}
\begin{tabular}{@{}clrccccccc@{}}
\toprule
\multirow{3}{*}{\textbf{\begin{tabular}[c]{@{}c@{}}Target\\      Language\end{tabular}}} & \multirow{3}{*}{\textbf{Method}} & \multirow{3}{*}{\textbf{Modality}} & \multicolumn{7}{c}{\textbf{BLEU}}                                                                             \\ \cmidrule(l){4-10} 
                                                                                         &                                  &                             & \multicolumn{6}{c}{\textbf{SNR}}                                                             & \textbf{clean} \\ \cmidrule(l){4-10} 
                                                                                         &                                  &                             & \textbf{-20 db} & \textbf{-10 db}  & \textbf{0 db}    & \textbf{10 db}   & \textbf{20 db}   & \textbf{Avg.} & \textbf{+$\infty$}    \\ \midrule
\multirow{4}{*}{En-Es}                                                              & Cascaded                        & A$_{\texttt{(+Noise)}}$            & 1.4$\pm$0.1          & 5.8$\pm$0.2          & 21.1$\pm$0.3          & 25.5$\pm$0.3          & 26.3$\pm$0.2          & 16.0          & 26.6                               \\
                                                                                    & AV-Hubert \cite{shi2022learning} & A$_{\texttt{(+Noise)}}$            & 1.5$\pm$0.2          & 6.7$\pm$0.2           & 22.3$\pm$0.4          & 27.7$\pm$0.2          & 28.6$\pm$0.3          & 17.6          & 28.9                               \\
                                                                                    & Cascaded                         & AV$_{\texttt{(+Noise)}}$           & 6.7$\pm$0.2          & 15.3$\pm$0.4          & 24.6$\pm$0.4          & 26.3$\pm$0.2          & 26.7$\pm$0.2          & 19.9          & 26.9                               \\
                                                                                    & AV-Hubert \cite{shi2022learning} & AV$_{\texttt{(+Noise)}}$           & \textbf{6.9$\pm$0.3} & \textbf{16.4\textbf{$\pm$0.5}} & \textbf{26.6$\pm$0.3} & \textbf{28.7$\pm$0.1} & \textbf{28.9$\pm$0.2} & \textbf{21.5} & \textbf{29.1}                      \\ \midrule
\multirow{4}{*}{En-Fr}                                                              & Cascaded                         & A$_{\texttt{(+Noise)}}$            & 1.3$\pm$0.2          & 4.5$\pm$0.3           & 16.6$\pm$0.4          & 20.9$\pm$0.3          & 21.3$\pm$0.1          & 12.9          & 21.7                               \\
                                                                                    & AV-Hubert \cite{shi2022learning} & A$_{\texttt{(+Noise)}}$            & 1.4$\pm$0.2          & 5.5$\pm$0.3           & 18.5$\pm$0.4          & 23.2$\pm$0.2          & 23.6$\pm$0.1          & 14.5          & 23.9                               \\
                                                                                    & Cascaded                         & AV$_{\texttt{(+Noise)}}$           & 4.6$\pm$0.1          & 11.4$\pm$0.5          & 19.4$\pm$0.3          & 21.5$\pm$0.2          & 22.0$\pm$0.2          & 15.8          & 22.3                               \\
                                                                                    & AV-Hubert \cite{shi2022learning} & AV$_{\texttt{(+Noise)}}$           & \textbf{4.9$\pm$0.2} & \textbf{12.1$\pm$0.3} & \textbf{21.6$\pm$0.4} & \textbf{23.7$\pm$0.3} & \textbf{24.3$\pm$0.1} & \textbf{17.3} & \textbf{24.6}                      \\ \midrule
\multirow{4}{*}{En-It}                                                              & Cascaded                         & A$_{\texttt{(+Noise)}}$            & 0.9$\pm$0.3          & 4.0$\pm$0.3           & 16.1$\pm$0.2          & 20.7$\pm$0.1          & 21.2$\pm$0.2          & 12.6          & 21.5                               \\
                                                                                    & AV-Hubert \cite{shi2022learning} & A$_{\texttt{(+Noise)}}$            & 1.0$\pm$0.2          & 5.1$\pm$0.5           & 18.3$\pm$0.3          & 22.7$\pm$0.2          & 23.6$\pm$0.2          & 14.1          & 23.8                               \\
                                                                                    & Cascaded                         & AV$_{\texttt{(+Noise)}}$           & 4.8$\pm$0.3          & 11.8$\pm$0.4          & 19.5$\pm$0.3          & 21.4$\pm$0.2          & 22.1$\pm$0.1          & 15.9          & 22.3                                   \\
                                                                                    & AV-Hubert \cite{shi2022learning} & AV$_{\texttt{(+Noise)}}$           & \textbf{5.0$\pm$0.4}          & \textbf{12.4$\pm$0.6}          & \textbf{21.9$\pm$0.3}          & \textbf{23.7$\pm$0.1}          & \textbf{24.1$\pm$0.2}          & \textbf{17.4}          & \textbf{24.5}                                   \\ \midrule
\multirow{4}{*}{En-Pt}                                                              & Cascaded                         & A$_{\texttt{(+Noise)}}$            & 1.1$\pm$0.3          & 5.4$\pm$0.5           & 20.1$\pm$0.4          & 24.9$\pm$0.2          & 26.0$\pm$0.1          & 15.5          & 26.2                                \\
                                                                                    & AV-Hubert \cite{shi2022learning} & A$_{\texttt{(+Noise)}}$            & 1.2$\pm$0.2          & 6.3$\pm$0.4           & 22.2$\pm$0.3          &  27.4$\pm$0.3          & 28.4$\pm$0.1          & 17.1          & 28.6                                   \\
                                                                                    & Cascaded                         & AV$_{\texttt{(+Noise)}}$           & 5.8$\pm$0.4          & 13.8$\pm$0.6          & 23.5$\pm$0.4          & 25.8$\pm$0.2          & 26.3$\pm$0.1          & 19.0          & 26.4                                   \\
                                                                                    & AV-Hubert \cite{shi2022learning} & AV$_{\texttt{(+Noise)}}$           & \textbf{6.1$\pm$0.3}          & \textbf{15.5$\pm$0.4}          & \textbf{26.0$\pm$0.3}              & \textbf{28.2$\pm$0.3}              & \textbf{28.6$\pm$0.2}          & \textbf{20.9}          & \textbf{28.8}                                   \\ \bottomrule

\end{tabular}
\end{center}
\caption{BLEU scores of audio speech translation and audio-visual speech translation with different noise SNRs.}
\label{NoisyST}
\end{table*}

\subsection{Quality of Translated Texts}
The translations in the AVMuST-TED dataset are taken directly from the high reliability translated subtitles in TED. TED has a very well-defined translation workflow to ensure that the translation accurately conveys the meaning, and we will now introduce it in detail. 
They recruit a total of 45,735 volunteers in 115 languages from all around the world, requiring each volunteer to be fluently bilingual in both source and target languages, fluent in the transcription language, and knowledgeable about what expressions are appropriate for subtitling. To ensure the quality of each assignment, each volunteer could apply for up to three editing assignments at the same time. Each volunteer can claim up to three editing assignments at a time to ensure the quality of each assignment. Each translation goes through three steps of transcription, translation and review before publishing, as shown in Figure \ref{fig:ted_trans}. TED provides an original transcript for all TED and TED-Ed content. For TEDx talks, volunteers are able to utilize auto-generated transcriptions as a base, or create their own from scratch. Subtitles are then translated from the original language into the target language, using a dynamic subtitle editor. Finally, before publication, subtitles are further reviewed by an experienced volunteer.
In Table \ref{tab:translation_text}, we present some sample translations of AVMuST-TED.

\section{Implementation Details}
\label{app:implement}
\paragraph{Audio and Visual Speeches Preprocessing.}
We follow the data preprocessing process in the prior work \cite{shi2022learning,afouras2018deep} for audio and visual speeches. For visual speech, we only extract the lip region as visual speech input, first detecting 68 facial keypoints using dlib \cite{king2009dlib}, and then aligning each face with the faces of its neighboring frames. From each visual speech utterance, we crop a $96\times96$ region-of-interest (ROI) lip-centered talking head video, representing the video speech. And for the audio speech, we also keep the same processing steps as the previous works \cite{shi2022learning,ma2021lira}. We extract 26-dimensional log filterbank energy feature from the raw waveform and stack 4 adjacent acoustic frames together for syncing with visual speech. we randomly crop a region of $88\times88$ from the entire ROI and perform a horizontal flip with probability 0.5 for data enhancement. we also apply noise with a probability of 0.25 to each audio utterance from \cite{Snyder2015MUSANAM} as steps in the prior works \cite{shi2022learning,afouras2018deep} for audio speech enhancement.

\paragraph{Training Details of MixSpeech.}
Our work is developed on the basis of the publicly available pre-trained model Transformer-Large of AV-Hubert \cite{shi2022learning}, which has 24 Transformer-LARGE with the embedding dimension/feed-forward dimension/attention heads of 1024/4096/16. Concretely, we adopt here the Transformer-LARGE model trained on LRS3 \cite{afouras2018lrs3} and VoxCeleb2 \cite{Chung2018VoxCeleb2DS}, augmented with noise. Correspondingly, for the translation decoder, we follow the same setup as AV-Hubert, with a 9-layer transformer decoder for easy comparison with it. During training, on one single 3090 GPU, we train 160K steps with labeled audio corpus, 80K of which are warmup steps; then we tune 40K steps with labeled visual corpus in the self-learning framework.

\section{Experiment}
\subsection{Speech Translation with Noise}
\label{app:noisy_speech}
In this section, we show the detailed performance of speech recognition in noisy environments in Table \ref{NoisyST}. Although the discrimination of audio speech is excellent and the performance of audio speech translation is outstanding, it is easily interfered by noise and  the performance of audio speech translation decreases rapidly with the enhancement of noise interference.
Following the previous works \cite{afouras2018deep,shi2022robust}, we add noise randomly sampled from MUSAN \cite{Snyder2015MUSANAM} to the audio speech and check the performance at five SNR levels \{-20, -10, 0, 10, 20\}db. For each experiment, we performed ten times, calculating the mean and the error to avoid interference from random sampling. 
The experimental results show that the performance of audio-visual speech translation is better than that of speech translation with audio speech only on all four languages in the noise-free environment (\ie, clear), demonstrating that visual speech further boosts the ceiling of speech recognition. 
Meanwhile, with the increase of noise interference (the smaller the SNR, the stronger the noise), the performance of audio speech translation decreases rapidly, especially during the process of SNR from 0db to -10db, the audio speech translation performance decreases most quickly, and the BLEU score decreases by -13.0 to -15.8. In contrast, speech translation with audio visual speech is significantly more resistant to noise, with the BLEU score decreasing by only -9.5 to -10.5 when SNR from 0db to -10db. At the same time, in terms of translation performance, all the audio-visual speech performances are better than those with only audio speech at the same SNR, and the audio-visual speech translation still performs well even at SNR = -10db, improving the robustness of the speech translation.

\begin{table*}[tb]
\tabcolsep=3pt
\begin{center}
\begin{tabular}{@{}l|cll@{}}
\toprule[1.5pt]
\multicolumn{4}{r}{\begin{minipage}[b]{\linewidth}\includegraphics[scale=0.82]{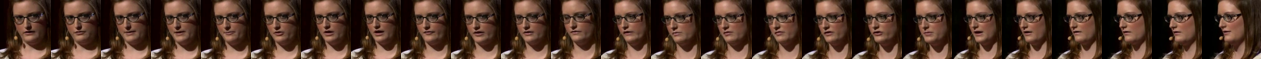}\end{minipage}} \\ \cmidrule(l){1-4} 
\multirow{6}{*}{\textbf{En-Es}}& \textbf{En}             & \small{TRXN:}       & that's why people often confuse me with a GPS.    \\ \cmidrule(l){2-4} 
                       & \multirow{5}{*}{\textbf{Es}}    & \small{GT:}       & por eso la gente me confunde a menudo con un gps    \\
                       &                                 & \small{A$_{\texttt{(+N)}}$:}     & \textcolor{lightgray}{por eso la gente} \textcolor{red}{\sout{ayúdame a lo que}} me \textcolor{lightgray}{confunde a menudo con un gps} \textcolor{red}{\sout{alegra por favor}}    \\
                       &                                 & \small{V:}     & por eso la gente a menudo me confunde con \textcolor{lightgray}{un gps} \textcolor{red}{\sout{los chimpancés}}    \\
                       &                                 & \small{A:}     & por eso la gente a menudo me confunde con \textcolor{lightgray}{un} \textcolor{blue}{(el)} gps    \\
                       &                                 & \small{AV:}     & por eso la gente a menudo me confunde con un gps    \\
\midrule[1.5pt]
 \multicolumn{4}{r}{\begin{minipage}[b]{\linewidth}\includegraphics[scale=0.82]{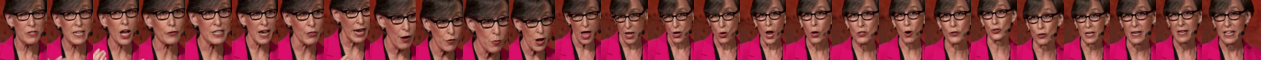}\end{minipage}} \\ \cmidrule(l){1-4} 
\multirow{6}{*}{\textbf{En-Fr}}& \textbf{En}             & \small{TRXN:}       & you need to understand that everyone who helps you on your journey    \\ \cmidrule(l){2-4} 
                       & \multirow{5}{*}{\textbf{Fr}}    & \small{GT:}       & vous devez comprendre que tous ceux qui vous aident durant votre voyage    \\
                       &                                 & \small{A$_{\texttt{(+N)}}$:}     & \textcolor{lightgray}{vous devez comprendre} que \textcolor{lightgray}{tous ceux qui vous} \textcolor{red}{\sout{avoir partagé avec un adolescent et}} \textcolor{lightgray}{aident} \textcolor{blue}{(aidé)} \textcolor{lightgray}{durant ...}   \\
                       &                                 & \small{V:}     & \textcolor{lightgray}{vous devez} \textcolor{red}{\sout{il faut}} comprendre que \textcolor{lightgray}{tous ceux} \textcolor{blue}{(chacun)} vous \textcolor{lightgray}{aident duran} \textcolor{blue}{(aide)} \textcolor{red}{\sout{à}} votre voyage    \\
                       &                                 & \small{A:}     & vous devez comprendre que \textcolor{lightgray}{tous ceux} \textcolor{red}{\sout{partout}} qui vous \textcolor{lightgray}{aident durant} \textcolor{blue}{(aide dans)} votre \textcolor{lightgray}{voyage} \textcolor{blue}{(parcours)}    \\
                       &                                 & \small{AV:}     & vous devez comprendre que \textcolor{lightgray}{tous ceux} \textcolor{blue}{(chaque personne)} qui vous \textcolor{lightgray}{aident durant} \textcolor{blue}{(aide dans)} votre voyage    \\

\midrule[1.5pt]
 \multicolumn{4}{r}{\begin{minipage}[b]{\linewidth}\includegraphics[scale=0.82]{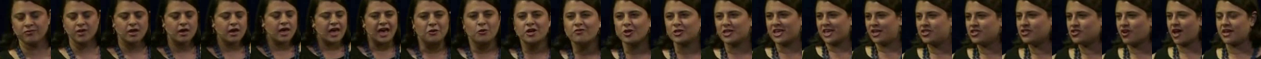}\end{minipage}} \\ \cmidrule(l){1-4} 
\multirow{6}{*}{\textbf{En-It}}& \textbf{En}             & \small{TRXN:}       & and one of our litigation strategies    \\ \cmidrule(l){2-4} 
                       & \multirow{5}{*}{\textbf{It}}    & \small{GT:}       & e una delle nostre strategie in tribunale    \\
                       &                                 & \small{A$_{\texttt{(+N)}}$:}     & \textcolor{lightgray}{e} \textcolor{red}{\sout{in}} una \textcolor{lightgray}{delle nostre strategie in tribunale} \textcolor{red}{\sout{di queste acque calde}}    \\
                       &                                 & \small{V:}     & e una delle nostre \textcolor{lightgray}{strategie in tribunale} \textcolor{red}{\sout{future eliminazioni}}    \\
                       &                                 & \small{A:}     & \textcolor{lightgray}{e} una delle nostre strategie \textcolor{lightgray}{in tribunale} \textcolor{red}{\sout{di contenzione}}    \\
                       &                                 & \small{AV:}     & e una delle nostre strategie \textcolor{lightgray}{in tribunale} \textcolor{red}{\sout{di}} \textcolor{blue}{(litigazione)}   \\
\midrule[1.5pt]
 \multicolumn{4}{r}{\begin{minipage}[b]{\linewidth}\includegraphics[scale=0.82]{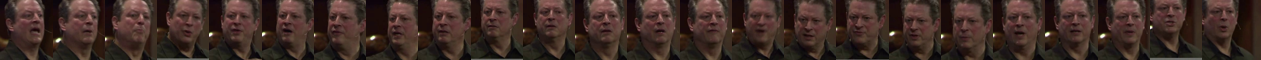}\end{minipage}} \\ \cmidrule(l){1-4} 
\multirow{6}{*}{\textbf{En-Pt}}& \textbf{En}             & \small{TRXN:}       & and both of the finalists for the Democratic nomination    \\ \cmidrule(l){2-4} 
                       & \multirow{5}{*}{\textbf{Pt}}    & \small{GT:}       & e ambos os finalistas para a nomeação democrática    \\
                       &                                 & \small{A$_{\texttt{(+N)}}$:}     & \textcolor{lightgray}{e ambos os finalistas } \textcolor{red}{\sout{tenho estado à espera de um minuto}} para \textcolor{red}{\sout{crescer no meio duma pessoa}} \textcolor{lightgray}{a ...}    \\
                       &                                 & \small{V:}     & e \textcolor{lightgray}{ambos} \textcolor{blue}{(ambas)} \textcolor{lightgray}{os finalistas} \textcolor{red}{\sout{as famílias democrática}} para a \textcolor{lightgray}{nomeação} democracia    \\
                       &                                 & \small{A:}     & e \textcolor{lightgray}{ambos} \textcolor{blue}{(os dois)} finalistas para a \textcolor{lightgray}{nomeação} \textcolor{red}{\sout{nação}} democrática    \\
                       &                                 & \small{AV:}    & e \textcolor{lightgray}{ambos} \textcolor{blue}{(os dois)} finalistas para a nomeação democrática    \\
\bottomrule[1.5pt]
\end{tabular}
\end{center}
\caption{Qualitative performance of the four target languages on the AVMuST-TED. Among them, A$_{(+\texttt{N})}$ for noisy audio in the SNR of -10db, V for visual, A for audio and AV for audio-visual. \textcolor{red}{\sout{Red Strikeout Words}}: mistranslated words with opposite meaning, \textcolor{blue}{(Blue Words in parentheses)}: mistranslated words with similar meaning, \textcolor{gray}{Gray Words}: the absent words. TRXN: transcript in English. GT: Ground Truth in the target language.}
\label{tab:visual_noise}
\end{table*}

\subsection{More Qualitative Analysis}
\label{app:qua}
To further quantitatively demonstrate the enhancement of visual speech to speech translation, we show more samples from AVMuST-TED and their outcomes with different modality speech translation in Table \ref{tab:visual_noise}.

\paragraph{Visual Speech VS Audio Speech with Noise}
Although the discrimination of visual speech is not as good as audio speech, it is not interfered by noise, and we choose the translation of audio speech in the SNR of -10db to compare with that of visual speech.

\paragraph{Audio-Visual Speech VS Audio Speech}
The robustness of speech translation can be further enhanced with the visual speech based on audio speech in the manner of audio-visual speech translation.

\section{Discussion}
\paragraph{Ethical Discussion}
Based on audio speech translation, visual speech for translation further enriches the application scenarios of speech translation technology (in silent or noise-bearing scenarios), while increasing the reliability of speech translation with the manner of audio-visual speech translation. As a cross-lingual translation technology, speech translation can be applied to many online applications (\eg, online medical, online education, \etc), contributing to the fairness of technology in disadvantaged areas. However, for visual speech, there could be some concerns about information leakage. But in fact, as we have mentioned before, lip reading and lip translation can only perform with high-definition, high-frame-rate frontal face videos that ensures clear visibility of lips and lip movements. Typically, only specially recorded videos, such as those from online meetings and public presentations, meet the strict video conditions that guarantee the unavailability of visual speech from videos such as surveillance for information leakage.

\paragraph{Limitations Discussion}
In this paper, we focus on the association between audio-visual speech and do not discuss the effect of machine translation datasets on lip translation yet. Many previous speech translation works have sufficiently demonstrated the enhancement of machine learning datasets for audio speech translation, and we have reasons to believe that it can also greatly improve the performance of lip translation, so there is no detailed discussion about it in this paper. Correspondingly, this paper focuses on a topic that has never appeared in other speech translation tasks, the interaction between audio-visual speech. Our follow-up work will address the blanks of this work.
\end{document}